%% file: main.tex
\documentclass[10pt,twocolumn,letterpaper]{article}

\usepackage{cvpr}              

\usepackage{graphicx}
\usepackage{textcomp}
\usepackage{gensymb}
\usepackage{amsmath}
\usepackage{amssymb}
\usepackage{booktabs}
\usepackage{threeparttable}
\usepackage[table,xcdraw]{xcolor}
\usepackage{float}
\restylefloat{table}
\usepackage{comment}
\usepackage{arydshln}
\usepackage{makecell}

\DeclareMathOperator*{\argmax}{arg\,max}
\let\svthefootnote\thefootnote
\newcommand\blankfootnote[1]{%
  \let\thefootnote\relax\footnotetext{#1}%
  \let\thefootnote\svthefootnote%
}

\usepackage[pagebackref,breaklinks,colorlinks]{hyperref}
\usepackage[super]{nth}

\usepackage[capitalize]{cleveref}
\crefname{section}{Sec.}{Secs.}
\Crefname{section}{Section}{Sections}
\Crefname{table}{Table}{Tables}
\crefname{table}{Tab.}{Tabs.}

\def\cvprPaperID{*****} 
\def\confName{CVPR}
\def\confYear{2022}

\begin{document}

\title{LidarMultiNet: Unifying LiDAR Semantic Segmentation, 3D Object Detection, and Panoptic Segmentation in a Single Multi-task Network}
\author{
 Dongqiangzi~Ye$^1$\footnotemark[1]~~~~~~~~~~Weijia~Chen$^1$\footnotemark[1]~~~~~~~~~~Zixiang~Zhou$^{1,2}$\footnotemark[1] \footnotemark[2]~~~~~~~~~~Yufei~Xie$^1$\footnotemark[1]\\
 Yu~Wang$^1$~~~~~~~~~~Panqu~Wang$^1$~~~~~~~~~~Hassan~Foroosh$^2$\\
 {$^1$TuSimple~~~$^2$University of Central Florida}\\
{\tt\small \{eowinye, weijia.chen619, zhouzixiang130, univerflyxie, yuwangrpi, wangpanqumanu\}@gmail.com} \\
{\tt\small Hassan.Foroosh@ucf.edu}
}

\maketitle

\renewcommand{\thefootnote}{\fnsymbol{footnote}}
\footnotetext[1]{Equal contribution}
\footnotetext[2]{Work done during an internship at TuSimple}

\begin{abstract}
   This technical report presents the \textbf{\nth{1}} place winning solution for the Waymo Open Dataset 3D semantic segmentation challenge 2022. Our network, termed \textbf{LidarMultiNet}, unifies the major LiDAR perception tasks such as 3D semantic segmentation, object detection, and panoptic segmentation in a single framework. At the core of LidarMultiNet is a strong 3D voxel-based encoder-decoder network with a novel Global Context Pooling (GCP) module extracting global contextual features from a LiDAR frame to complement its local features. An optional second stage is proposed to refine the first-stage segmentation or generate accurate panoptic segmentation results. Our solution achieves a mIoU of 71.13 and is the best for most of the 22 classes on the Waymo 3D semantic segmentation test set, outperforming all the other 3D semantic segmentation methods on the official leaderboard ~\cite{waymoseg2022}. We demonstrate for the first time that major LiDAR perception tasks can be unified in a single strong network that can be trained end-to-end.
\end{abstract}

\input{introduction}
\input{method}

\input{experiment}

\section{Conclusion}
Our proposed LidarMultiNet reached the \nth{1} place in the Waymo Open Dataset 3D semantic segmentation challenge 2022. As one of our future works, we plan to verify the effectiveness of LidarMultiNet on 3D object detection and panoptic segmentation benchmarks as well.

{\small
\bibliographystyle{ieee_fullname}
\bibliography{egbib}
}

\end{document}

%% file: introduction.tex
\section{Introduction}
LiDAR 3D semantic segmentation is a fundamental perception task of autonomous driving. With the recent release of several large-scale LiDAR point cloud datasets with semantic labels~\cite{behley2019iccv,caesar2020nuscenes,sun2020scalability}, more methods are proposed to advance the research in LiDAR semantic segmentation.

\begin{figure}
    \centering
    \begin{subfigure}[b]{0.49\linewidth}
         \frame{\includegraphics[width=\textwidth]{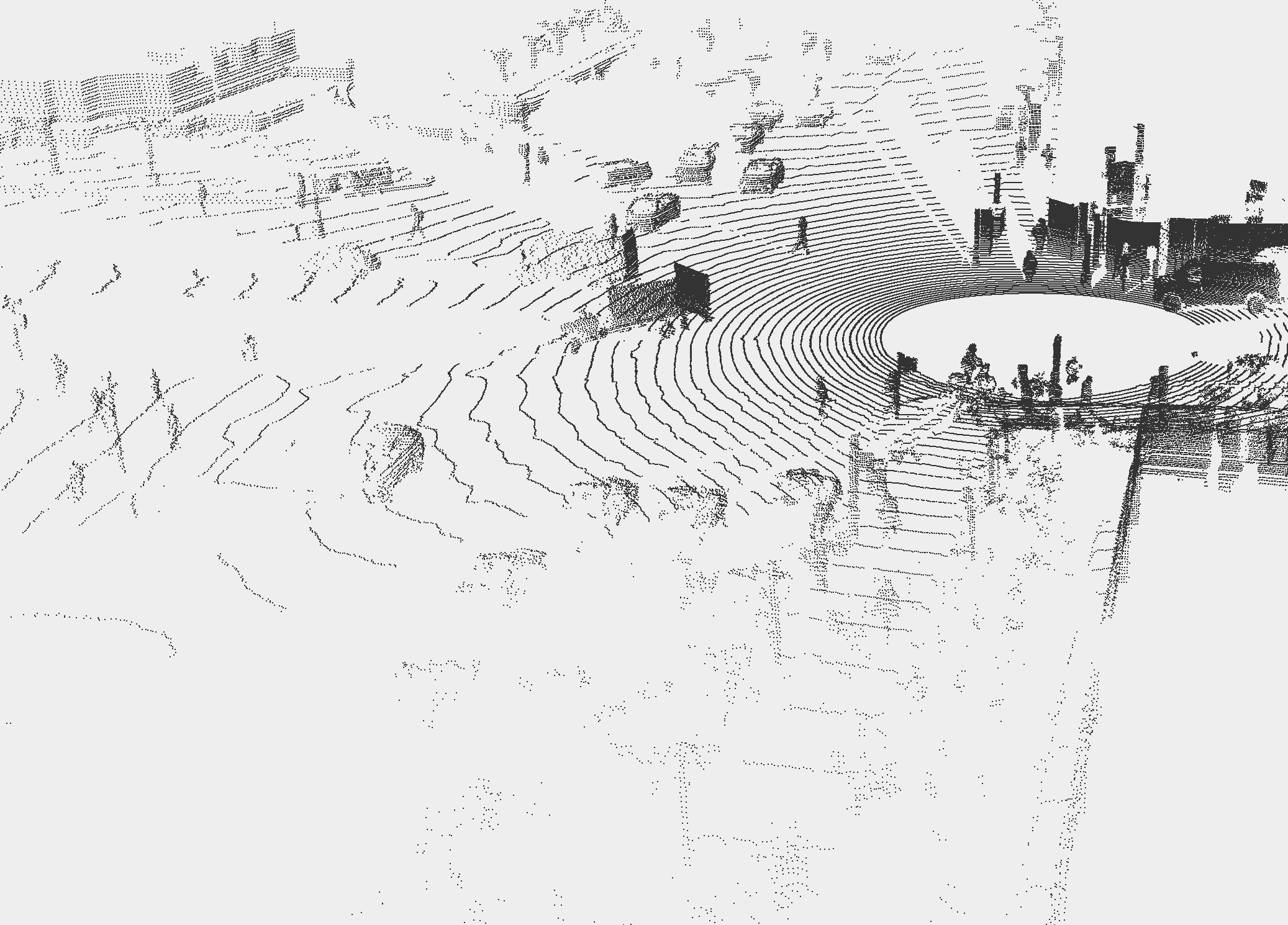}}
        \caption{}
    \end{subfigure}
    \begin{subfigure}[b]{0.49\linewidth}
        \frame{\includegraphics[width=\textwidth]{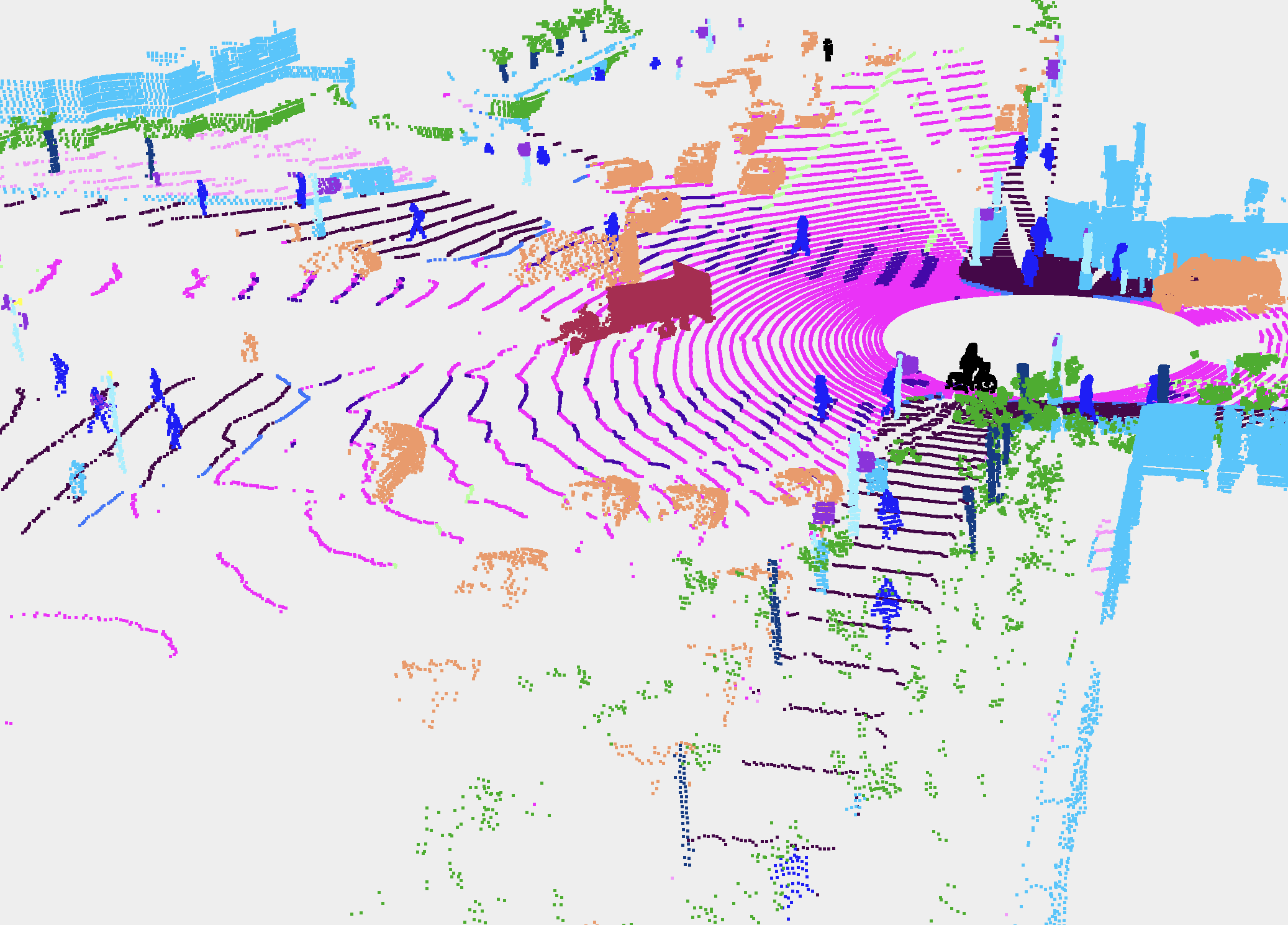}}
        \caption{}
    \end{subfigure}
    \begin{subfigure}[b]{0.49\linewidth}
        \frame{\includegraphics[width=\textwidth]{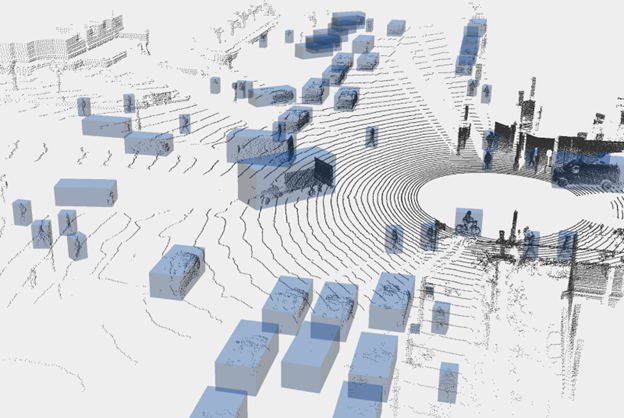}}
        \caption{}
    \end{subfigure}
    \begin{subfigure}[b]{0.49\linewidth}
        \frame{\includegraphics[width=\textwidth]{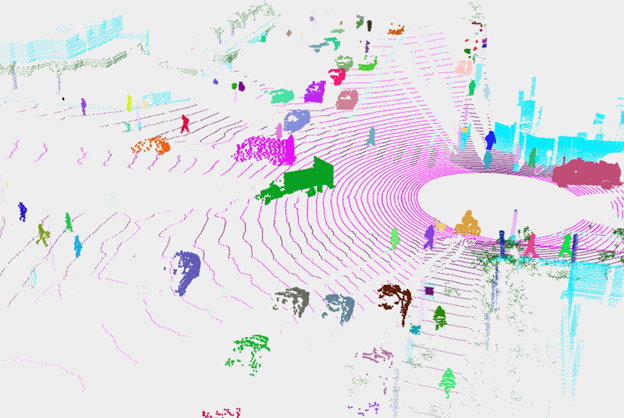}}
        \caption{}
    \end{subfigure}
    \caption{Our LidarMultiNet takes LiDAR point cloud (a) as input and performs simultaneous 3D semantic segmentation (b), 3D object detection (c) and panoptic segmentation (d) in a single unified network.}
    \label{fig:example}
\end{figure}

Compared to 2D image and 3D indoor point cloud segmentation, outdoor LiDAR point cloud presents more challenges for the segmentation problem. Due to the large-scale and sparsity of LiDAR point clouds, well-studied 2D and indoor 3D semantic segmentation methods~\cite{qi2017pointnet,qi2017pointnet++} fail to directly adjust to LiDAR semantic segmentation. With the advent of sparse convolution~\cite{yan2018second,choy20194d}, more methods have started to segment the point clouds in the 3D voxel space. However, due to the requirement of sparse convolution and the need for an encoder-decoder structure for the segmentation task, previous voxel-based LiDAR segmentation networks~\cite{zhu2021cylindrical, Cheng_2021_CVPR} have difficulty to learn global contextual information. On the other hand, more recent works try to fuse features from multiple views that contain both voxel-level and point-level information. These approaches focus more on exploiting local point geometrical relations to recover fine-grained details for segmentation.

Major LiDAR perception tasks including 3D segmentation (\eg, PolarNet~\cite{Zhang_2020_CVPR}), 3D object detection (\eg, CenterPoint~\cite{yin2021center}) and panoptic segmentation (\eg, Panoptic-PolarNet ~\cite{Zhou2021CVPR}), are usually performed in separate and independent networks. In this work, we propose to unify these three major LiDAR perception tasks in a single network that exploits the synergy between these tasks and achieves state-of-the-art performance, as shown in Figure~\ref{fig:example}.

Our main contributions are summarized as follows:

\begin{itemize}
    \item We present an efficient voxel-based LiDAR multi-task network that unifies the major LiDAR perception tasks.
    \item We propose a novel Global Context Pooling (GCP) module to improve the global feature learning in the encoder-decoder architecture based on 3D sparse convolution.
    \item We introduce a second-stage refinement module to refine the first-stage semantic segmentation or produce accurate panoptic segmentation results.
\end{itemize}

\section{Related Work}
\noindent\textbf{LiDAR Semantic Segmentation}
LiDAR point cloud semantic segmentation usually requires transforming the large-scale sparse point cloud into either a 3D voxel map, 2D Bird's Eye View (BEV), or range-view map. Following the trend in the LiDAR point cloud detection methods~\cite{yang2018pixor,lang2019pointpillars}, PolarNet~\cite{Zhang_2020_CVPR} projected the point cloud into a 2D polar BEV map to balance the point distribution in the voxelization procedure. Thanks to the development of the 3D sparse convolution layer~\cite{yan2018second,choy20194d} in point cloud processing, more methods~\cite{zhu2021cylindrical, Cheng_2021_CVPR} start using 3D sparse CNN on a voxel feature map for the LiDAR point cloud segmentation. 
In contrast to the recent methods~\cite{zhu2021cylindrical,Cheng_2021_CVPR,ye2021drinet,xu2021rpvnet} that focus on the fine-grained features for details, our method aims to enhance the global feature learning in the voxel-based segmentation network.

\noindent\textbf{Two-stage Semantic Segmentation}
Two-stage or multi-stage refinement is a common procedure in point cloud object detection networks~\cite{shi2020pv,yin2021center,li2021lidar,Sheng2021ICCV}, but is rarely used in point cloud semantic segmentation. In the image domain, various methods~\cite{li2016iterative,zhu20183d,chen20182,zhang2019acfnet,yuan2020object} use multiple stages to refine the segmentation prediction from coarse to fine. 

\noindent\textbf{LiDAR Panoptic Segmentation}
Recent LiDAR panoptic segmentation methods~\cite{Zhou2021CVPR,hong2021lidar,razani2021gp} usually derive from the well-studied segmentation networks~\cite{Zhang_2020_CVPR,zhu2021cylindrical,Cheng_2021_CVPR} in a bottom-up fashion. This is largely due to the loss of height information in the detection networks, which makes them difficult to adjust the learned feature representation to the segmentation task. This results in two incompatible designs for the best segmentation~\cite{xu2021rpvnet} and detection~\cite{yin2021center} methods. According to ~\cite{fong2021panoptic}, end-to-end LiDAR panoptic segmentation methods still underperform compared to independently combined detection and segmentation models. In this work, our model can perform simultaneous 3D object detection and semantic segmentation and trains the tasks jointly in an end-to-end fashion.

\noindent\textbf{Multi-Task Network}
MultiNet~\cite{multinet2016} is a seminal work of the image-based multi-task learning that unifies object detection and road understanding tasks in a single network. In LiDAR-based perception, LidarMTL~\cite{lidarmtl2021} proposed a simple and efficient multi-task network based on 3D sparse convolution and deconvolutions for joint object detection and road understanding. 

%% file: method.tex
\section{Method}

\begin{figure*}
\centering
\includegraphics[width=\textwidth]{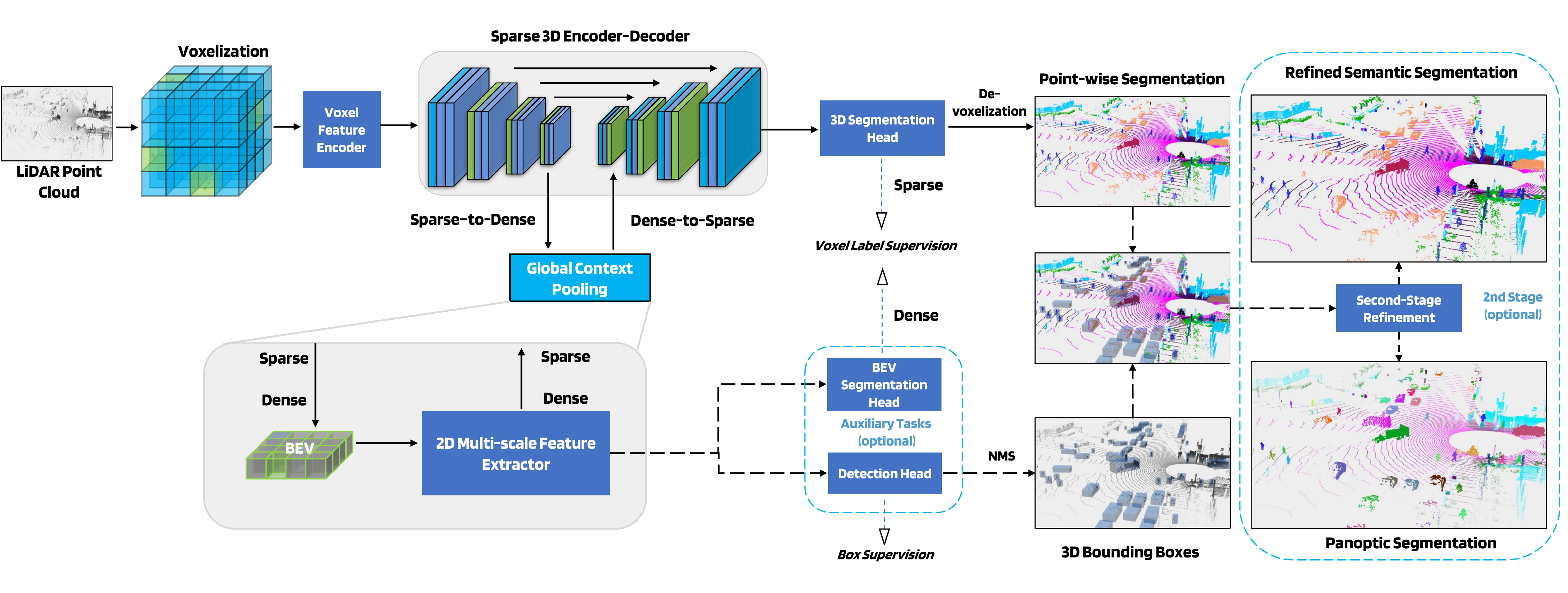}
\caption{\textbf{Main Architecture of the LidarMultiNet}. At the core of our network is a 3D encoder-decoder based on the 3D sparse convolution and deconvolutions. In between the encoder and the decoder, we apply our novel Global Context Pooling (GCP) module to extract contextual information through the conversion between sparse and dense feature maps and via a 2D multi-scale feature extractor. The 3D segmentation head is attached to the decoder output and its predicted voxel labels are projected back to the point-level via a de-voxelization step. Two auxiliary heads, namely the BEV segmentation head and 3D detection head can be attached to the 2D BEV feature map. An optional 2nd-stage produces refined semantic segmentation and panoptic segmentation results.}
\label{fig:architecture}
\end{figure*}

\subsection{Voxel-based LiDAR Segmentation}
LiDAR point cloud semantic segmentation aims to predict semantic labels $L=\{l_{i}|l_{i}\in (1\dots K)\}_{i=1}^{N}$ for all points $P=\{p_{i}|p_{i}\in\mathbb{R}^{3+c}\}_{i=1}^{N}$ in the point cloud, where $N$ denotes the total number of points, $K$ is the number of classes and each point has $(3+c)$ input features, \ie, the 3D coordinates $(x,y,z)$, intensity of the reflection, LiDAR elongation, timestamp, etc. We want to learn a segmentation model $f$, parameterized by $\theta$, that minimizes the difference between the prediction $L = f(P| \theta)$ and the point-wise semantic annotation $L^{gt}=\{l_{i}^{gt}|l_{i}^{gt}\in (1\dots K)\}_{i=1}^{N}$.

\noindent\textbf{Voxelization}
We first transform the point cloud coordinates $(x,y,z)$ into the voxel index $\{v_{i}=(\lfloor \frac{x_{i}}{s_{x}} \rfloor,\lfloor\frac{y_{i}}{s_{y}}\rfloor, \lfloor \frac{z_{i}}{s_{z}}\rfloor)\}_{i=1}^{N}$, where $s$ is the voxel size. Then, we use a simple voxel feature encoder, which only contains Multi-Layer Perceptron (MLP) and maxpooling layers to generate the sparse voxel feature representation $\mathcal{V}\in\mathbb{R}^{M\times C}$:

\begin{equation}
   \mathcal{V}_{j} = \max\limits_{v_{i}=v_{j}}(\textnormal{MLP}(p_{i})) , j\in (1\dots M)
\end{equation}
where $M$ is the number of unique voxel indices. We also generate the ground truth label of each sparse voxel through majority voting: $L_{j}^{v}=\argmax\limits_{v_{i}=v_{j}}(l_{i}^{gt})$.

\noindent\textbf{Sparse Voxel-based Encoder-decoder Backbone Network}
We designed a 3D counterpart of the 2D UNet~\cite{ronneberger2015u} as the backbone of our segmentation network, where the voxel feature is gradually downsampled to $\tfrac{1}{8}$ of the original size in the encoder and upsampled back in the decoder.
Key indices are shared among the encoder and decoder in the same scale to keep the same sparsity and reduce repetitive calculation of sparse indices. In each decoder block, the voxel features from previous block and lateral skip-connected features from the encoder are concatenated and further processed by a residual submanifold convolution block and then upsampled by inverse sparse convolution. We supervise our model with the voxel-level label $L^{v}$ during training and project the predicted label back to the point-level via a de-voxelization step during inference.

\begin{figure}
  \begin{center}
    \includegraphics[width=0.25\textwidth]{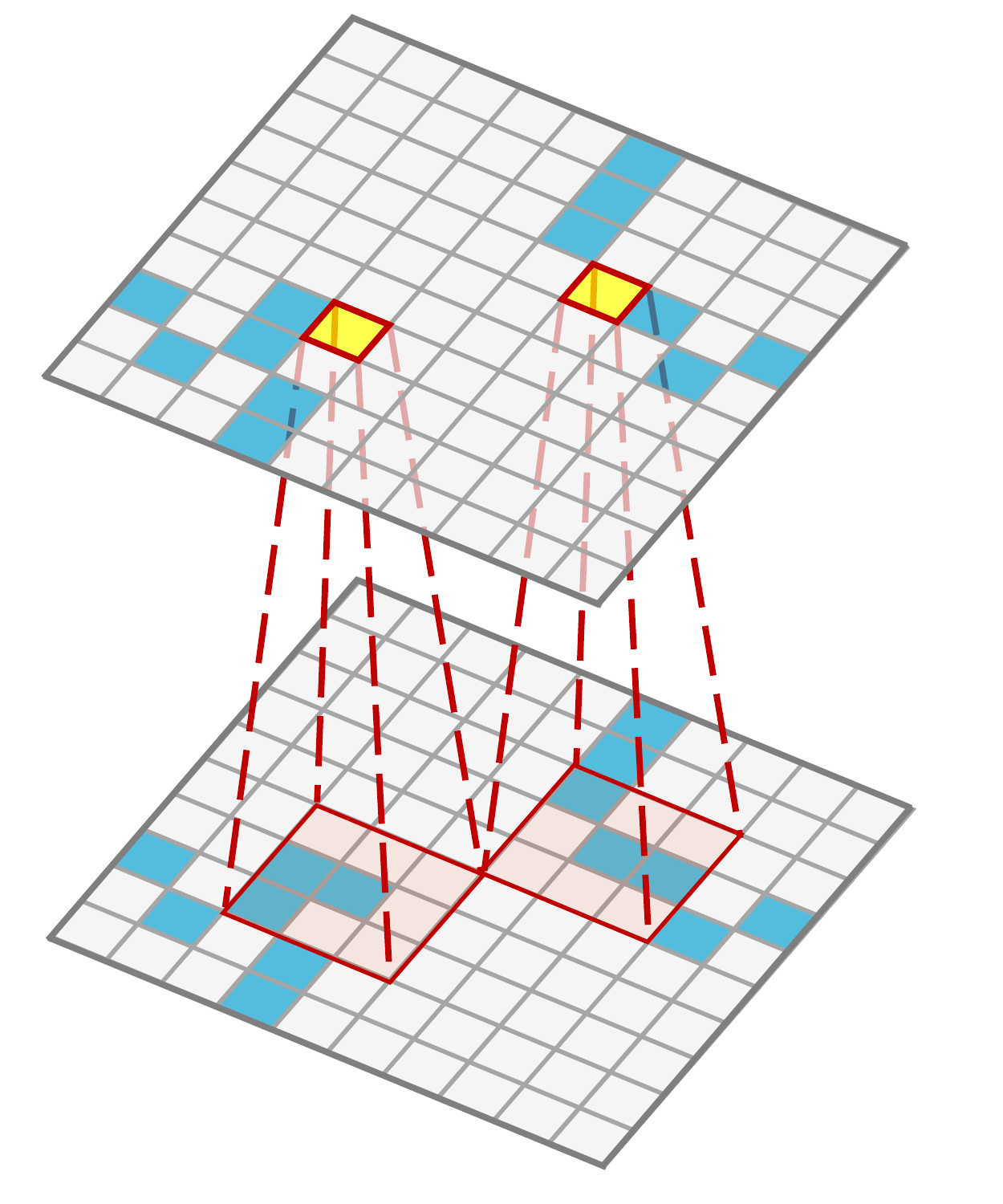}
  \end{center}
  \caption{\textbf{Submanifold Sparse Convolution.} Convolution only takes place at valid voxels. Features cannot reach the voxels that have big gaps in between.}
  \label{fig:subm}
\end{figure}

\subsection{LidarMultiNet}
The main architecture of LidarMultiNet is illustrated in Figure~\ref{fig:architecture}. The main network consists of an encoder-decoder architecture based on 3D sparse convolution and deconvolutions. The novel GCP module is applied in between the encoder and the decoder. Two auxiliary tasks, namely 3D object detection and BEV segmentation can be attached to the network. An optional second stage can be applied to refine semantic segmentation and generate panoptic segmentation results. 

\subsection{Global Context Pooling}
3D sparse convolution drastically reduces the memory consumption and runtime of the 3D CNN, but it generally requires the layers of the same scale to retain the same sparsity in both encoder and decoder. This restricts the network to using only submanifold convolution~\cite{graham20183d} (Figure~\ref{fig:subm}) in the same scale. However, submanifold convolution cannot broadcast features to isolated voxels through stacking multiple convolution layers, limiting the ability to learn global contextual information. Global Context Pooling (GCP) is designed to extract global contextual feature from 3D sparse tensor by projecting it to 2D dense BEV feature map and then converting the dense feature map back after applying 2D multi-scale feature extractor. 

\begin{figure}
  \begin{center}
    \includegraphics[width=0.48\textwidth]{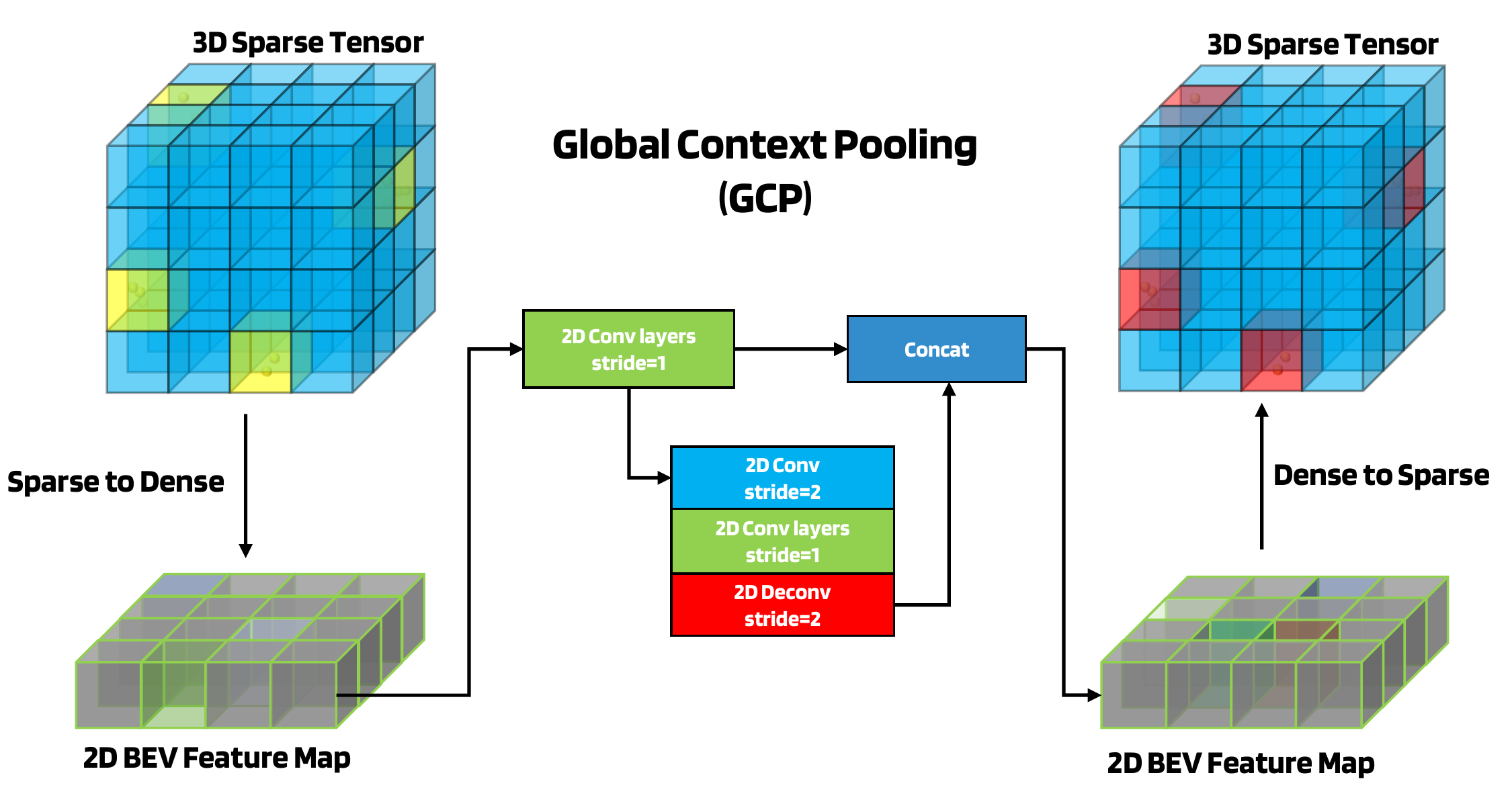}
  \end{center}
  \caption{\textbf{Illustration of the Global Context Pooling (GCP) module}. 3D sparse tensor is projected to a 2D BEV feature map. Two levels of 2D BEV feature maps are concatenated and then converted back to a 3D sparse tensor.}
  \label{fig:gcp}
\end{figure}

As illustrated in Figure~\ref{fig:gcp}, given the 3D sparse tensor output by the encoder, we first project it into a dense BEV feature map $\mathcal{F}^{sparse}\in\mathbb{R}^{C\times M'} \to \mathcal{F}^{dense}\in\mathbb{R}^{C\times \frac{D}{d_{z}} \times \frac{H}{d_{x}} \times \frac{W}{d_{y}}}$, where $d$ is the downsampling ratio and $M'$ is the number of valid voxels in the last scale. We concatenate the features across the height dimension together to form a 2D BEV feature map $\mathcal{F}^{bev}\in\mathbb{R}^{(C*\frac{D}{d_{z}}) \times \frac{H}{d_{x}} \times \frac{W}{d_{y}}}$. Then, we use a 2D multi-scale CNN to further extract contextual information. Lastly, we reshape the encoded BEV feature representation to the dense voxel map, then transform it back to the sparse voxel feature following the reverse dense-to-sparse conversion.

Benefiting from GCP, our architecture could significantly enlarge the receptive field, which plays an important role in semantic segmentation. In addition, the BEV feature maps in GCP can be shared with other tasks (\eg, object detection) simply by attaching additional heads with slight increase of computational cost. By utilizing the BEV-level training like object detection, GCP can enhance the segmentation performance furthermore.

\begin{figure*}
\centering
\includegraphics[width=0.9\textwidth]{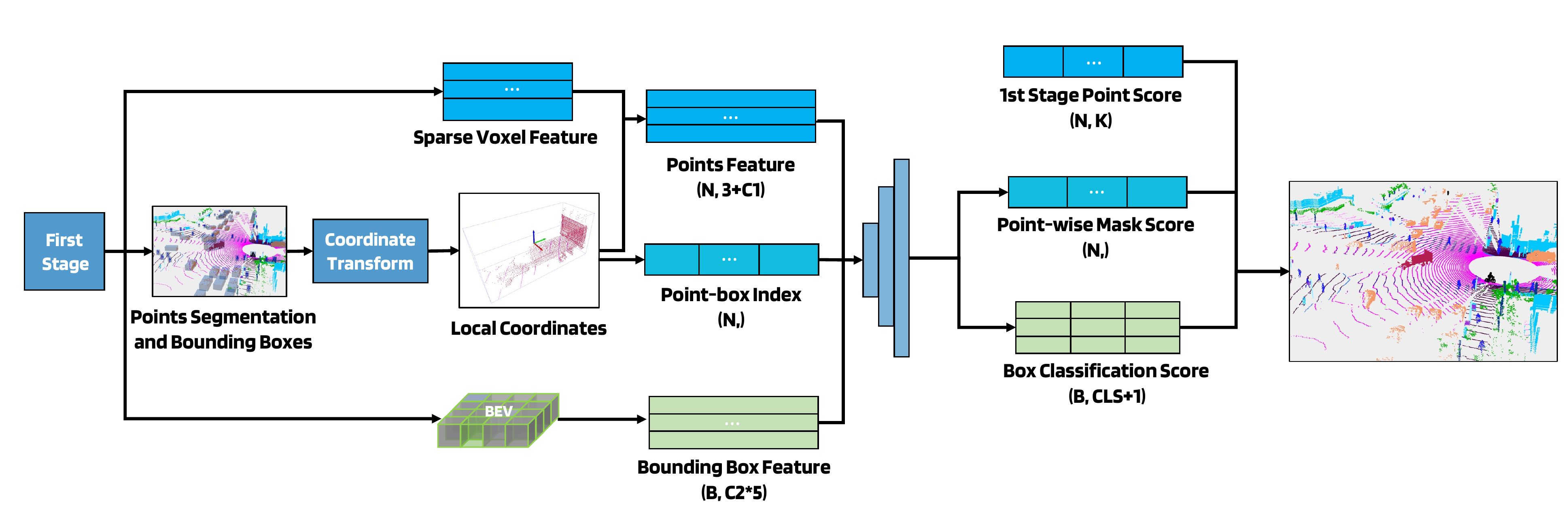}
\caption{\textbf{Illustration of the second-stage refinement pipeline.} The architecture of the second-stage refinement is point-based. We first fuse the detected boxes, voxel-wise features, and BEV features from the 1st stage to generate the inputs for the 2nd stage. We apply the local coordinate transformation to the points within each box. Then, a point-based backbone with MLPs, attention modules, and aggregation modules infers the box-wise classification scores and point-wise mask scores. The final scores are computed by fusing the 1st and 2nd stage predictions.}
\label{fig:2nd_stage_refinement_illustration}
\end{figure*}

\subsection{LiDAR Multi-task Learning}
\noindent\textbf{Auxiliary Tasks}
We attach the detection head of CenterPoint~\cite{yin2021center}, which is widely used in anchor-free 3D detection, to the multi-scale feature extractor. The motivations are twofold. First, the detection loss can serve as the auxiliary loss and help exploit the synergy between the segmentation and detection tasks, since Waymo Open Dataset (WOD) has both segmentation and bounding box labels for the thing classes (\ie, countable objects such as pedestrians and vehicles). Second, the network transcends into a multi-task network that performs both tasks simultaneously.

Besides the detection head, an additional BEV segmentation head also can be attached to the 2D branch of the network, providing coarse segmentation results and serving as another auxiliary loss during training.  

\noindent\textbf{Multi-task Training and Losses} 
The 3D segmentation branch predicts voxel-level label $L^{v}=\{l_{j}|l_{j}\in (1\dots K)\}_{j=1}^{M}$ given the learned voxel features $\mathcal{F}^{sparse}\in\mathbb{R}^{C_{voxel}\times M}$ output by the 3D decoder. We supervise it through a combination of cross-entropy loss and Lovasz loss~\cite{berman2018lovasz}: $\mathcal{L}_{SEG} = \mathcal{L}_{ce}^{v} + \mathcal{L}_{Lovasz}^{v}$. Note that $\mathcal{L}_{SEG}$ is of sparse loss and the computational cost as well as GPU memory usage is much less than dense loss.

The detection head is applied on the 2D BEV feature map: $\mathcal{F}^{bev}\in\mathbb{R}^{C_{bev} \times \frac{H}{d_{x}} \times \frac{W}{d_{y}}}$. It predicts a class-specific heatmap, object size and orientation, and IoU rectification score, which are supervised by the focal loss~\cite{lin2017focal} ($\mathcal{L}_{hm}$) and L1 loss ($\mathcal{L}_{reg}, \mathcal{L}_{iou}$) respectively: $\mathcal{L}_{DET} = \lambda_{hm}\mathcal{L}_{hm}+\lambda_{reg}\mathcal{L}_{reg}+\lambda_{iou}\mathcal{L}_{iou}$, where the weights $\lambda_{hm}$, $\lambda_{reg}$, $\lambda_{iou}$ are set to $[1,2,1]$.

The BEV segmentation head is supervised with $\mathcal{L}_{BEV}$, a dense loss consisting of cross-entropy loss and Lovasz loss: $\mathcal{L}_{BEV} = \mathcal{L}_{ce}^{bev}+\mathcal{L}_{Lovasz}^{bev}$.

Our network is trained end-to-end with supervision by $\mathcal{L}_{SEG}$, $\mathcal{L}_{DET}$ and $\mathcal{L}_{BEV}$, in which both $\mathcal{L}_{DET}$ and $\mathcal{L}_{BEV}$ serve as auxiliary losses during training. We apply a weighted multi-task loss \cite{kendall2018multi} $\mathcal{L}_{total}$ defined as
 \begin{equation}
\mathcal{L}_{total} =\sum_{i\in \left \{SEG,DET,BEV  \right \}
}\frac{1}{2\sigma^{2}_{i} }\mathcal{L}_{i}+\frac{1}{2} \log\sigma^{2}_{i} 
\end{equation}
 where  $\mathcal\sigma_{i}$ is the learned parameter representing the uncertainty degree of $task_{i}$. The more uncertain the $task_{i}$ is, the less $\mathcal{L}_{i}$ contributes to $\mathcal{L}_{total}$. The second term can be treated as a regulator for $\mathcal\sigma_{i}$ in training.

\subsection{Second-stage Refinement}
\begin{figure*}
\centering
\includegraphics[width=0.9\textwidth]{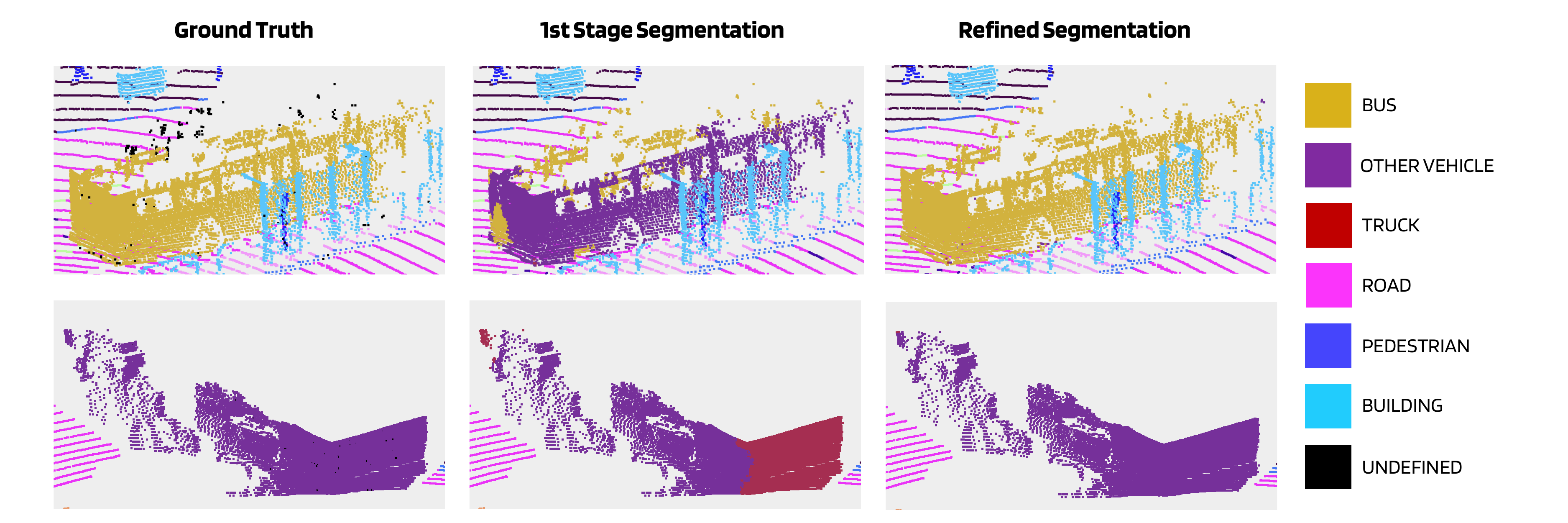}
\caption{\textbf{Examples of the 2nd-stage refinement}. The segmentation consistency of points of the thing objects can be improved by the 2nd stage.}
\label{fig:2nd_stage_refinement_example}
\end{figure*}

During training the detection serves as an auxiliary loss of the segmentation network, not contributing to the segmentation directly. As shown in Figure~\ref{fig:2nd_stage_refinement_example}, the points within a detection bounding box might be misclassified as multiple classes due to the lack of spatial prior knowledge. In order to improve the spatial consistency for the thing classes, we propose a novel point-based approach as the second stage which can also be used to provide accurate panoptic segmentation results. The second stage is illustrated in Figure~\ref{fig:2nd_stage_refinement_illustration}.

Specifically, given all points $P=\{p_{i}|p_{i}\in\mathbb{R}^{3+c}\}_{i=1}^{N}$ in the point cloud, the $B$ predicted bounding boxes, sparse voxel features of the final segmentation branch $\mathcal{F}^{sparse}$, and BEV feature maps $\mathcal{F}^{bev}$, our 2nd stage predicts box-wise class scores $S_{box}$ and point-wise mask scores $S_{point}$. We first transform each point within bounding boxes into its local coordinate as $P_{local}$. Then we devoxelize all points to find their voxel features and process 2nd stage points-wise features as $P_{2nd} = concat(P_{local}, P_{voxel})$. We extract 2nd stage box-wise features $\mathcal{B}_{2nd}$ as in \cite{yin2021center} from BEV feature maps $\mathcal{F}^{bev}$. For box-wise aggregation, we generate point-box index $I = \{{ind}_{i}|{ind}_{i}\in\mathbb{I}, 0 \leq {ind}_{i} \leq B\}_{i=1}^{N}$. Note that stuff classes points are assigned with index $\varnothing$ and they will not be refined in the second stage. Next, our point-wise backbone takes $P_{2nd}$, $I$ and $\mathcal{B}_{2nd}$ as inputs and outputs point-wise mask scores ${S}_{point} = \{{sp}_{i}|{sp}_{i}\in (0, 1)\}_{i=1}^{N}$ and box-wise classification score ${S}_{box} = \{{sb}_{i}|{sb}_{i}\in {(0, 1)}^{cls + 1}\}_{i=1}^{B}$, where $cls$ denotes the number of thing classes and the one additional class is the unrefined class $\emptyset$. During training, we supervise box-wise class scores through cross-entropy loss and point-wise mask scores through binary cross-entropy loss.

We compute the final scores by fusing the 1st-stage prediction and the 2nd-stage prediction. Moreover, we can integrate the refined segmentation results with the predicted box scores to perform panoptic segmentation.

%% file: experiment.tex
\definecolor{Gray}{gray}{0.95}
\newcolumntype{g}{>{\columncolor{Gray}}c}

\section{Experiments}

\subsection{Waymo Open Dataset}

The Waymo Open Dataset \cite{sun2020scalability} contains 1150 sequences in total, split into 798 in the training set, 202 in the validation set, and 150 in the test set. Each sequence contains about 200 frames and the 3D Semantic Segmentation labels are provided for sampled frames. We use the v1.3.2 dataset which contains 23,691 and 5,976 frames with semantic segmentation labels in the training set and validation set, respectively. There are a total of 2,982 frames in the final test set. Waymo Open Dataset has semantic labels for a total of 23 classes, including an undefined class.

\begin{table}
    \centering
    \begin{tabular}{lcc}
    \Xhline{4\arrayrulewidth}
        Hyperparameters & Module & Values \\ \hline
        \#Input features & VFE & 16\\
        Voxel size & Voxelization & $(0.1m, 0.1m, 0.15m)$\\
        Range $x/y$ & Voxelization & $[-75.2m, 75.2m]$ \\
        Range $z$ & Voxelization & $[-2m, 4m]$ \\
        Downsampling & Encoder & 8 \\
        3D Depth & Encoder & 2, 3, 3, 3 \\ 
        3D Width & Encoder & 32, 64, 128, 256 \\ 
        3D Width & Decoder & 128, 64, 32, 32 \\ 
        2D Depth & GCP & 6, 6  \\
        2D Width & GCP &  128, 256 \\
        \Xhline{4\arrayrulewidth}
    \end{tabular}
    \caption{Implementation details of the LidarMultiNet}
    \label{table:design_space}
    \vspace{-8pt}
\end{table}

\begin{table*}
\centering
\resizebox{\linewidth}{!}{
\begin{threeparttable}
\begin{tabular}{l|l|llllllllllllllllllllll}
\Xhline{4\arrayrulewidth}
\textbf{Waymo Leaderboard}                   & mIoU   & \rotatebox{80}{CAR}    & \rotatebox{80}{TRUCK}  & \rotatebox{80}{BUS}    & \rotatebox{80}{OTHERVEHICLE} & \rotatebox{80}{MOTORCYCLIST} & \rotatebox{80}{BICYCLIST} & \rotatebox{80}{PEDESTRIAN} & \rotatebox{80}{SIGN}   & \rotatebox{80}{TRAFFICLIGHT} & \rotatebox{80}{POLE}   & \rotatebox{80}{CONE} & \rotatebox{80}{BICYCLE} & \rotatebox{80}{MOTORCYCLE} & \rotatebox{80}{BUILDING} & \rotatebox{80}{VEGETATION} & \rotatebox{80}{TREETRUNK} & \rotatebox{80}{CURB}   & \rotatebox{80}{ROAD}   & \rotatebox{80}{LANEMARKER} & \rotatebox{80}{OTHERGROUND} & \rotatebox{80}{WALKABLE} & \rotatebox{80}{SIDEWALK} \\ \hline
\textbf{LidarMultiNet}             & \textbf{71.13} & \textbf{95.86} & \textbf{70.57} & 81.44 & 35.49         & 0.77       & \textbf{90.66}    & \textbf{93.23}     & \textbf{73.84} & \textbf{33.14}         & \textbf{81.37} & 64.68             & 69.92  & 76.73     & \textbf{97.37}   & \textbf{88.92}     & \textbf{73.58}      & \textbf{76.54} & 93.21 & \textbf{50.37}       & \textbf{53.82}        & \textbf{75.48}   & \textbf{87.95}   \\ 
SegNet3DV2       & 70.48 & 95.73 & 69.03 & 79.74 & \textbf{37.0}           & 0            & 88.77    & 92.66     & 71.82 & 30.02         & 80.85 & \textbf{65.97}             & 69.53  & \textbf{76.97}     & 97.15   & 88.18     & 72.76      & 76.4  & \textbf{93.27} & 49.49       & 52.61        & 75.4    & 87.25   \\ 
HorizonSegExpert & 69.44 & 95.55 & 68.93 & \textbf{84.38} & 29.39         & 0.95       & 86.06    & 92.22     & 72.71 & 32.27         & 80.56 & 55.67             & 68.56  & 71.7      & 97.18   & 87.81     & 72.49      & 76.0   & 93.09 & 48.91       & 52.21        & 73.78   & 87.29   \\ 
HRI\_HZ\_SMRPV   & 69.38 & 95.79 & 66.97 & 78.36 & 32.8          & 0.04       & 89.06    & 91.83     & 72.93 & 31.73         & 79.82 & 61.33             & \textbf{70.23}  & 76.02     & 96.94   & 87.2      & 72.15      & 73.8  & 92.6  & 48.02       & 50.82        & 71.94   & 86.03   \\ 
Waymo\_3DSEG     & 68.99 & 95.47 & 69.54 & 77.85 & 31.29         & 0.67       & 84.4     & 91.12     & 72.73 & 31.45         & 80.12 & 61.25             & 68.87  & 74.87     & 96.96   & 87.89     & 72.93      & 74.62 & 92.61 & 44.68       & 49.33        & 73.1    & 86.13   \\ 
SPVCNN++         & 67.7  & 95.12 & 67.73 & 75.61 & 35.16         & 0            & 85.1     & 91.57     & 73.24 & 31.94         & 78.85 & 61.4              & 65.97  & 73.6      & 90.82   & 86.58     & 70.54      & 75.53 & 91.73 & 41.31       & 40.22        & 71.5    & 85.87   \\ 
PolarFuse        & 67.28 & 95.06 & 67.79 & 77.04 & 30.54         & 0            & 79.08    & 89.61     & 65.37 & 30.38         & 78.53 & 57.3              & 62.97  & 67.96     & 96.38   & 87.07     & 70.26      & 75.1  & 92.54 & 48.91       & 51.75        & 71.39   & 85.02   \\ 
LeapNet          & 66.89 & 94.45 & 65.68 & 79.01 & 30.74         & 0.01       & 79.44    & 90.07     & 70.46 & 30.34         & 77.67 & 52.71             & 61.58  & 66.84     & 96.65   & 87.01     & 69.8       & 74.49 & 92.61 & 44.91       & 47.17        & 73.59   & 86.34   \\ 
3DSEG                     & 66.77 & 94.64 & 66.95 & 77.61 & 30.67         & \textbf{2.11}       & 78.59    & 89.18     & 70.01 & 31.0           & 77.4  & 56.73             & 60.81  & 67.91     & 96.55   & 87.33     & 70.89      & 70.92 & 91.79 & 42.99       & 46.78        & 72.95   & 85.04   \\ 
CAVPercep                 & 63.73 & 93.64 & 62.83 & 68.12 & 22.83         & 1.89       & 71.73    & 87.39     & 67.03 & 29.41         & 74.29 & 55.03             & 55.6   & 60.84     & 96.15   & 86.48     & 68.06      & 68.33 & 91.24 & 41.26       & 46.19        & 70.55   & 83.15   \\ 
VS-Concord3D              & 63.54 & 92.6  & 66.9  & 73.13 & 24.26         & 0.83       & 76.77    & 85.51     & 66.88 & 29.98         & 75.81 & 53.86             & 62.26  & 68.11     & 86.04   & 75.13     & 66.95      & 67.95 & 90.6  & 40.43       & 45.43        & 65.6    & 82.96   \\ 
RGBV-RP Net               & 62.61 & 94.87 & 67.48 & 74.91 & 33.09         & 0            & 77.34    & 88.66     & 68.05 & 28.68         & 74.78 & 37.69             & 53.82  & 64.92     & 96.59   & 86.47     & 67.18      & 70.93 & 90.97 & 23.7        & 24.01        & 68.89   & 84.48  \\ 
\Xhline{4\arrayrulewidth}
\end{tabular}
\end{threeparttable}
}
\caption{Waymo Open Dataset Semantic Segmentation Leaderboard. Our \textbf{LidarMultiNet} reached the highest mIoU of 71.13 and achieved the best accuracy on 15 out of the 22 classes.}
\label{table:leaderboard}
\end{table*}

\subsection{Implementation Details}

As summarized in Table~\ref{table:design_space}, the 3D encoder in our network consists of 4 stages of 3D sparse convolutions with increasing channel width 32, 64, 128, 256. Each stage starts with a sparse convolution layer followed by two submanifold sparse convolution blocks. The first sparse convolution layer has a stride of 2 except at the first stage. Therefore the spatial resolution is downsampled by 8 times in the encoder. The 3D decoder has 4 stages of 3D sparse deconvolution blocks with decreasing channel widths 128, 64, 32, 32. 2D dense BEV features are obtained by projecting the 3D sparse tensor via channel concatenation across its height dimension. 

During training, the point cloud range is set to $[-75.2m, 75.2m]$ for $x$ axis and $y$ axis, and $[-2m, 4m]$ for $z$ axis, and the voxel size is set to $(0.1m, 0.1m, 0.15m)$. For data augmentation, we use random flipping with respect to $xz$-plane and $yz$-plane, and global scaling with a random factor from $[0.95, 1.05]$. We also apply a random global rotation between [$-\pi/4, \pi/4$] and a random translation with a normally distributed displacement $\Delta d \sim \mathcal{N}(0, 0.5^{2})$ along all axes. 

Following \cite{yin2021center}, we transform the past two LiDAR frames using the vehicle's pose information and merge them with the current LiDAR frame to produce a denser point cloud input and append a timestamp feature to each LiDAR point. Points of past LiDAR frames participate in the voxel feature computation but do not contribute to the loss calculation.

We train the models using AdamW \cite{loshchilov2017decoupled} optimizer with one-cycle learning rate policy \cite{sylvain2018onecycle}, with max learning rate 3e-3, weight decay 0.01, and momentum 0.95 to 0.85. We use a batch size of 2 on each of the 8 V100 GPUs. For one-stage model, we train the models from scratch for 20 epochs. For the two-stage model, we freeze the 1st stage and finetune the 2nd stage for 6 epochs.

\begin{table}
    \centering
    \resizebox{\linewidth}{!}{
    \begin{tabular}{ccccccc}
    \Xhline{4\arrayrulewidth}
    Baseline & Multi-frame & GCP & $\mathcal{L}_{BEV}$ & $\mathcal{L}_{DET}$ & Two-Stage & mIoU                           \\\hline
    $\checkmark$ & & & & & & 69.90   \\
    $\checkmark$ & $\checkmark$ & & & & & 70.49 \\
    $\checkmark$ & $\checkmark$ & $\checkmark$ & & & & 71.43 \\
    $\checkmark$ & $\checkmark$ & $\checkmark$ &$\checkmark$ & & & 71.58 \\
    $\checkmark$ & $\checkmark$ & $\checkmark$ &$\checkmark$ &$\checkmark$ & &72.06 \\
    $\checkmark$ & $\checkmark$ & $\checkmark$ &$\checkmark$ & $\checkmark$ &$\checkmark$ &72.40 \\
    \Xhline{4\arrayrulewidth}
    \end{tabular}
    }
    \caption{Ablation studies for 3D semantic segmentation on the WOD validation set.We show the improvement introduced by each component compared to our LidarMultiNet base network.}
    \label{table:ablation}
    \vspace{-8pt}
\end{table}

\subsection{Ablation Study} 
We ablate each component of the LidarMultiNet and the result on the WOD validation set is shown in Table~\ref{table:ablation}. Our base LidarMultiNet reaches a mIoU of 69.90 on the validation set. On top of this baseline, multi-sweep input (\ie, including past two frames) brings a 0.59 mIoU improvement. Our proposed GCP further improves the mIoU by 0.94. The auxiliary losses (\ie, BEV segmentation and 3D object detection) result in a total improvement of 0.63 mIoU, and the 2nd-stage further improves the mIoU by 0.34, forming our best single model on the WOD validation set. 

\subsection{Test-Time Augmentation and Ensemble} \label{ssec:tta_ensemble}

In order to further improve the performance, we apply Test-Time Augmentation (TTA) and Ensemble methods.
Specifically, we use flipping with respect to $xz$-plane and $yz$-plane, [$0.95$, $1.05$] for global scaling, and [$\pm22.5\degree$, $\pm45\degree$, $\pm135\degree$, $\pm157.5\degree$, $180\degree$] for yaw rotation. Besides, we found pitch and roll rotation were helpful in the segmentation task, and we use $\pm8\degree$ for pitch rotation and $\pm5\degree$ for roll rotation. In addition, we also apply $\pm0.2m$ translation along $z$-axis for augmentation.

Besides the best single model, we also explored the network design space and designed multiple variants for model ensemble. For example, more models are trained with smaller voxel size $(0.05m, 0.05m, 0.15m)$, smaller downsample factor ($4\times$), different channel width (64), without the 2nd stage, or with more past frames (4 sweeps). Table \ref{table:tta_ensemble} shows the improvement brought by TTA and ensemble on the validation set. For our submission to the leaderboard, a total of 7 models are ensembled to generate the segmentation result on the test set. 

\begin{table}
    \centering
    \begin{tabular}{l|c}
    \Xhline{4\arrayrulewidth}
        Methods & mIoU \\ \hline
        Best single model & 72.40 \\
        + TTA & 73.05    \\
        + Ensemble & 73.78 \\
    \Xhline{4\arrayrulewidth}
    \end{tabular}
    \caption{Improvements made on top of the best single model on the WOD validation set}
    \label{table:tta_ensemble}
    \vspace{-8pt}
\end{table}

\subsection{Waymo Open Dataset 3D Semantic Segmentation Challenge Leaderboard} 
For our final submission to the leaderboard, we train our models from scratch on the joint dataset of Waymo training and validation splits. Table~\ref{table:leaderboard} is the final WOD semantic segmentation leaderboard and shows that our LidarMultiNet achieves a mIoU of 71.13 and ranks the \nth{1} place on the leaderboard, and also has the best IoU for 15 out of the total 22 classes.